\documentclass[12pt, a4paper]{article}

\usepackage[a4paper,margin=20mm]{geometry}
\usepackage[utf8]{inputenc}
\usepackage{times}
\usepackage{color} 
\usepackage{caption}
\usepackage{siunitx}
\usepackage{gensymb}
\usepackage{float}
\usepackage{graphicx}
\usepackage{amssymb}
\usepackage{tabularx}

\begin{document}


\newcommand{\HRule}{\rule{\linewidth}{0.5mm}} 
\begin{center}

\HRule \\[0.4cm]
{ \huge \bfseries Artificial SA-I and RA-I Afferents\\ for Tactile Sensing of Ridges and Gratings}\\[0.4cm] 
\HRule \\[1.5cm]

\textbf{Nicholas~Pestell, Thom Griffith and~Nathan~F.~Lepora}\\[0.5cm]
Department of Engineering Mathematics and Bristol Robotics Laboratory,\\ University of Bristol, BS8 1QU, UK\\
e-mail: n.lepora@bristol.ac.uk\\[0.5cm]
{\large \today}\\[3cm]
\begin{abstract}
For robot touch to reach the capabilities of human touch, artificial tactile sensors may require transduction principles like those of natural tactile afferents. Here we propose that a biomimetic tactile sensor (the TacTip) could provide suitable artificial analogues of the tactile skin dynamics, afferent responses and population encoding. Our 3D-printed sensor skin is based on the physiology of the dermal-epidermal interface with an underlying mesh of biomimetic intermediate ridges and dermal papillae, comprising inner pins tipped with markers. Slowly-adapting SA-I activity is modelled by marker displacements and rapidly-adapting RA-I activity by marker speeds. We test the biological plausibility of these artificial population codes with three classic experiments used for natural touch: (1a) responses to normal pressure to test adaptation of single afferents and spatial modulation across the population; (1b) responses to bars, edges and gratings to compare with measurements from monkey primary afferents; and (2) discrimination of grating orientation to compare with human perceptual performance. Our results show a match between artificial and natural touch at single afferent, population and perceptual levels. As expected, natural skin is more sensitive, which raises a challenge to fabricate a biomimetic fingertip that demonstrates human sensitivity using the transduction principles of human touch.
\end{abstract}
\end{center}
  
\pagebreak

\section{Introduction}

The fields of neuroscience and robotics are converging as machines approach animals in their capabilities \cite{prescott_living_2018}. Within the somatosensory modalities, a natural convergence is around the engineering of biologically-plausible artificial tactile sensory systems \cite{dahiya_tactile_2010,lucarotti_synthetic_2013}. A biomimetic tactile fingertip or skin can inform about the neurophysiology of human touch \cite{scheibert_role_2009,platkiewicz,delhaye}, lead to improved contact sensing for dexterous robots \cite{kappassov,ward-cherrier_tactip_2018,boutry_hierarchically_2018} and enable neuroprosthetics to restore a sense of touch in amputees \cite{oddo_intraneural_2016,george}. A fundamental question for all these application is how are stimuli represented (coded) in the peripheral nervous system? To answer this question for the sense of touch, one needs to consider the dynamics of the local skin tissue to which individual mechanoreceptive units respond, how individual afferent nerve fibres represent these dynamics when stimulated, and how the spatial aspects of tactile stimulation are represented across the population of tactile afferents. 

Here we propose that a biomimetic optical tactile sensor called the TacTip \cite{chorley1, ward-cherrier_tactip_2018} provides a suitable artificial analogue for these three aspects: skin dynamics, afferent response and spatial population encoding. Our proposition is underpinned by two hypotheses regarding peripheral neural codes in natural touch. First, the spike rates of individual natural afferents commonly provide a suitable metric to model artificial tactile channels because of their correlation with external stimuli \cite{brette}. There are several examples of experimental studies of touch that support this position~ \cite{johnson,phillips2,connor,srinivasan2}. Second, population codes, more specifically the spatial patterns of firing rates across afferent populations, are the primary abstract coding scheme for spatially-detailed static stimuli in natural touch. Again, there is a large body of experimental studies that support this position~\cite{srinivasan,goodwin,srinivasan2,johnson}.   

\begin{figure}[t]
    \centering
	\includegraphics[width=\linewidth]{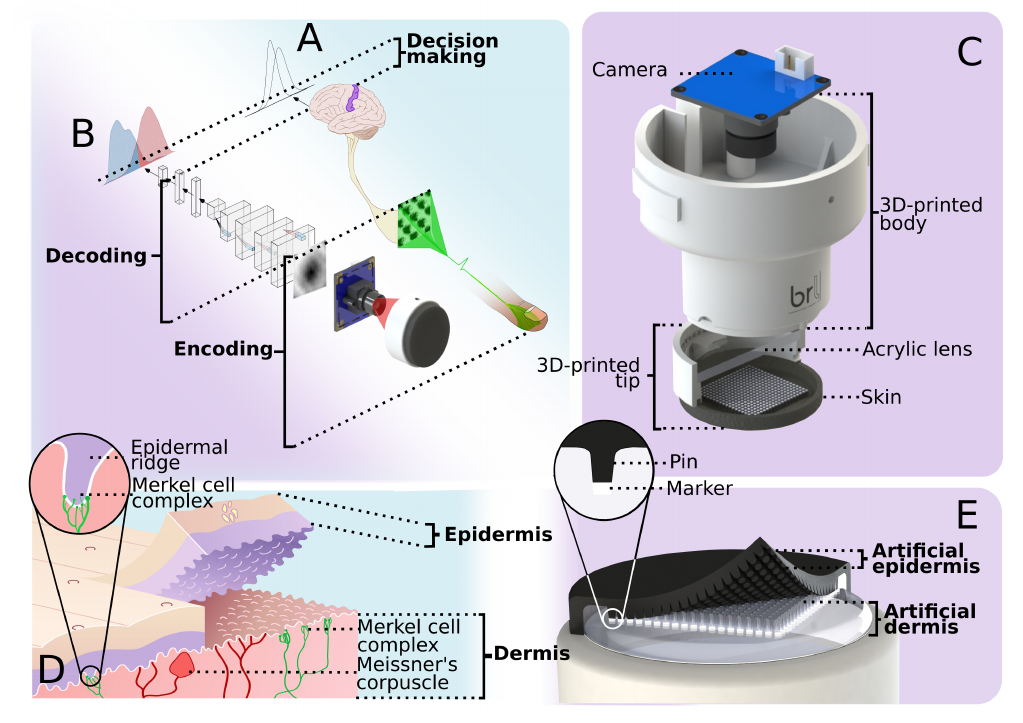}
	\caption{(\textbf{A},\textbf{B}) Perceptual pipelines of human and robot touch. In human touch, decoding is performed in the somatosensory cortex (highlighted). Here, in robot touch, an artificial decoder is constructed from a convolutional neural network. (\textbf{C}) Exploded view of the BRL TacTip showing its 3D-printed body and soft tip. (\textbf{D}) Diagram of human glabrous skin showing the stiff outer epidermis with epidermal ridges protruding into the softer inner dermis. Inset: Merkel cell complexes (SA-I mechanoreceptors) are located at the tips of epidermal ridges. (\textbf{E})~Diagram of the TacTip with stiff artificial epidermis over a softer artificial dermis (clear silicone gel). Natural epidermal ridges are replicated with stiff pins protruding into the silicone gel. Inset: the pins are tipped with white markers for artificial mechanoreception.}
	\label{bioT}
\end{figure}

Human glabrous skin has an intricate morphology of layers, microstructures and sensory receptors that underlie its many functions, from protecting the body to sensing surface contact~\cite{klatzky_touch_2003,abraira,zimmerman_gentle_2014}. The TacTip design seeks to mimic the shallow dermal and epidermal outer layers of human skin~\cite{chorley1,ward-cherrier_tactip_2018}. It has an outer biomimetic epidermis made from a rubber-like material over a soft inner biomimetic dermis made from elastomer gel (Figure~\ref{bioT}). These two materials interdigitate in a mesh of biomimetic intermediate ridges and dermal papillae, comprising stiff inner nodular pins that extend under the biomimetic epidermis into the soft gel. This structure mechanically amplifies the surface deformation of skin into a lateral movement of markers on the pin tips, analogous to a hypothesised function of the dermal papillae in human skin. Markers are tracked optically, with raw tactile data of marker displacements directly showing the shear-strain field within the biomimetic dermal-epidermal interface.

We introduce two novel feature sets derived from tracking the markers in the video output of the TacTip optical tactile sensor, which are intended to mimic the slowly-adapting SA-I and rapidly-adapting RA-I afferents of human touch. Natural SA-I firing rates are modelled by the absolute marker displacements from their at-rest positions and natural RA-I firing rates are modelled by the marker speeds. The rationale behind these models is that marker displacement is sustained with a static stimulus and marker velocity is non-zero only when the stimulus changes. The markers form a 22$\times$22\,mm$^2$ square array comprising 361 artificial SA-I and RA-I afferents with a density of $\sim$70\,cm\textsuperscript{-2}, about half the innervation density of type-I afferents in the human fingertip \cite{johansson2,johansson3}. 

These proposals are examined with three experiments adopted from the experimental study of human touch. Experiment 1a studies the response of artificial afferents to normal force stimulation on a flat plate. Experiment 1b studies the afferent reponse to complex aperiodic grating stimuli comprising edges, gaps and ridges. Experiment 2 studies the perception of grating orientation using the population codes from these artificial afferents. The latter two experiments mirror two classic studies of human and primate tactile population coding by Phillips and Johnson (1981) \cite{johnson,phillips2}. In addition to gaining insight into the capacity of artificial type-I afferents to mediate viable population codes, some novel methodologies are introduced for the testing and quantification of robot touch using well-established techniques from psychology such as signal detection theory and psychometric curves. Overall, a close match is found between artificial and natural touch in the responses of single SA-I and RA-I afferents, their population coding capabilities and the resulting perceptual performance.  

\section{Methods}
\label{sec04}

\subsection{Sensor Design}
\label{sec04:01:01}

Here we describe the key operating principles of the TacTip biomimetic optical tactile sensor and its customization to the sensor presented here. For manufacturing and detailed explanations of the design concepts, we refer the reader to a 2018 paper on `The TacTip Family' \cite{ward-cherrier_tactip_2018} and a recent (2021) review of `Soft Biomimetic Optical Tactile Sensing with the TacTip' \cite{lepora2021}.

The TacTip (Figure \ref{bioT}C) features a black flexible skin covering a clear silicone gel (Figure \ref{bioT}E). The artificial epidermis and dermis provide contrasting stiffness which is proposed to mimic the corresponding layers of natural skin. The artificial epidermis is 3D-printed as a flat surface, but once the tip is filled with gel it forms a slight convex bulge owing to pressure within the tip (Figure \ref{models}A).

In the human fingertip, Merkel cell complexes (SA-I afferents) are situated in the dermis at the tips of stiff epidermal ridges that protrude into the softer dermis (Figure \ref{bioT}D). This morphology is mimicked in the TacTip with stiff epidermal pins that protrude into the soft gel. A white marker on the tip of each pin provides an optical signature of local shear strain imaged using a USB camera (Figure \ref{bioT}C). Although the use of optics is clearly non-biomimetic, we consider the skin morphology and transduction of contact as based on the physiology of human skin.  361 white markers arranged in a 19$\times$19 square grid have a marker density of $\sim$70\,cm\textsuperscript{-2} ($\sim$1.2\,mm separation), which is approximately half the innervation density of type-I afferents in the human fingertip \cite{johansson2,johansson3}. 

\subsection{Sensor Fabrication}

The sensor body is fully 3D-printed in ABS thermoplastic using a Fortus 450mc (Stratasys) (Figure~1C). A USB camera (ELP-USBFHD01M-L21) is fixed to the inside of the body and used with resolution 640$\times$480 pixels at 90\,fps. The body features a bayonet mounting system for fixing onto a robot arm. A PCB ring sits close at the interface with the tip, with 6 white LEDs that illuminate the markers for imaging by the camera.

The tip is 3D-printed using an Objet 260 Connex multi-material 3D-printer (Stratasys), with the artificial epidermis, pins, markers and hard rim mechanically fused and printed as a single part (Figure 1C). The flexible skin and pins are printed in a black, rubber-like material (Tango Black+; Shore A 26-28), with the rim and markers in a rigid white plastic (VeroWhite, Stratasys). A circular cap is laser cut from clear acrylic and glued into the rim creating a cavity between the skin and cap. The cavity is manually injection filled with a clear silicone gel (RTV27905, Techsil; ShoreOO 10). The gel provides a stiffness to the tip which helps minimise hysteresis whilst maintaining compliance.

\subsection{Feature Extraction}
\label{sec04:02}

For deriving artificial afferent response from each marker, the marker positions $(x,y)$ in pixels are tracked frame-by-frame with a blob detection algorithm implemented in Python OpenCV. These markers are treated as counterparts to the Merkell cell and Meissner's corpuscle mechanoreceptors of type-I tactile afferents (Figure~1D).

\subsubsection{Artificial SA-I Afferents}
\label{sec04:02:01}

To account for the tonic nature of Merkel cell complexes, SA-I firing is modelled as marker displacements (shear-strain magnitude). Thus the SA-I response, SA$_{n,i}$, at frame $i$ for marker (afferent) $n$ is computed as the Euclidean distance between those marker positions and an initial rest frame ($i=0$):
\begin{equation}
{\rm SA}_{n,i} = \sqrt{(x_{n,i}-x_{n,0})^{2}+(y_{n,i}-y_{n,0})^{2}},
\label{SA-fire}
\end{equation}
where $1\leq n\leq N_{\rm markers}=361$ represents each of the artificial afferents.

\subsubsection{Artificial RA-I Afferents}
\label{sec04:02:02}

To account for the phasic nature of Meissner's corpuscles, RA-I firing is modelled as marker speed. This model is similar to other proposed transduction models of RA-I firing, where the first derivative of pressure is used as the primary input to a biological neuron model \cite{lee,bensmaia6,kim2}. Thus the RA-I response, RA$_{n,i}$, at frame $i$ for marker (afferent) $n$ is computed as the absolute difference between the SA-I responses for adjacent frames: 
\begin{equation}
{\rm RA}_{n,i} = |{\rm SA}_{n,i} - {\rm SA}_{n,i-1}|.
\label{RA-I-fire}
\end{equation}
These two equations are simplifications of the afferent activity and anatomy; for example, Meissner's corpuscles and Merkel cells are found at different locations across the dermal-epidermal boundary and biological SA-I activity is not relative to an initial fixed frame. Overall, our intent is to provide a parsimonious account of afferent activity from simple transformations of the optical tactile output, which can be built upon in future work to improve the biological realism.

\begin{figure}[t]
	\begin{center}
		\includegraphics[width=.9\linewidth]{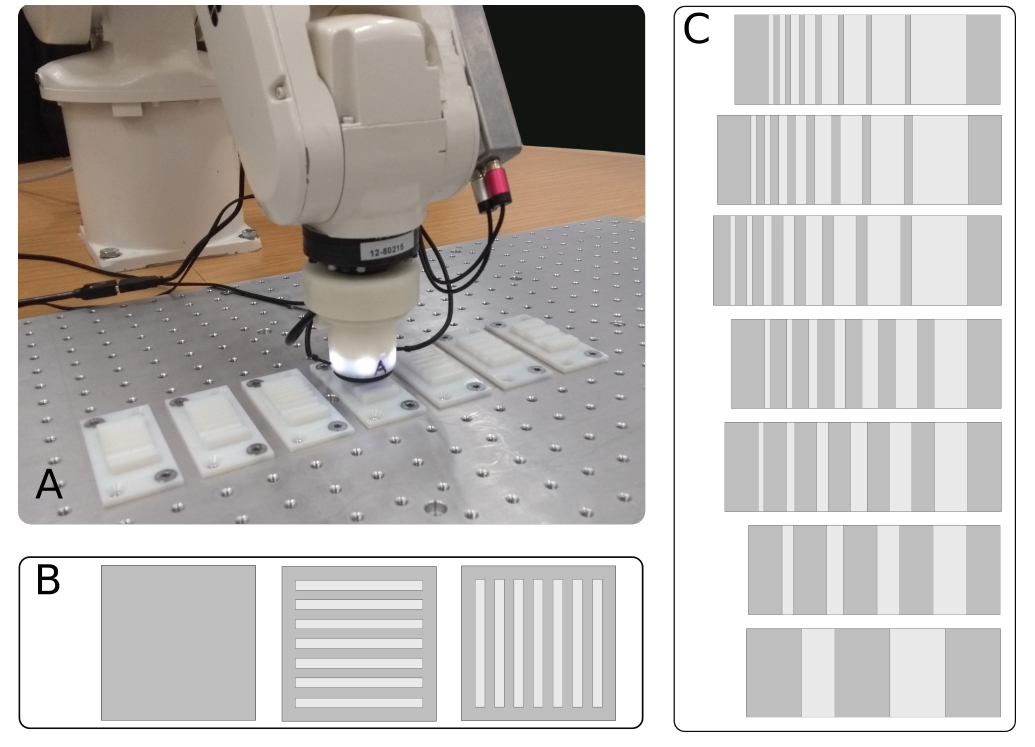}
	\end{center}
	\caption{Robot setup and stimuli. \textbf{(A)} TacTip mounted as an end effector to the industrial robot arm, collecting data. \textbf{(B)} Plan view of flat plate stimulus (Experiment 1a) and two orientations of a periodic grating stimulus (Experiment 2). \textbf{(C)} Plan view of seven aperiodic gratings (Experiment 1b).}
	\label{robot_gratings}
\end{figure}

\subsection{Experiments}

The same robot system is used for all experiments (Figure~\ref{robot_gratings}A), comprising the TacTip soft biomimetic optical tactile sensor mounted as the end effector on a 6-axis industrial robot arm (IRB120, ABB). The arm is mounted adjacent to an aluminium plate for attaching the grating stimuli. 

\subsubsection{Experiment 1a: Response to Normal Pressure}

{\bf Collection Procedure:} The TacTip tactile sensor begins in free-space at 2\,mm above a flat 3D-printed surface (Figure \ref{robot_gratings}B, left). Tactile image data is recorded as the sensor is moved downward until a compression of 2.5\,mm is achieved, where the sensor is held stationary for 3\,seconds before returning to its initial position where the recording stops. To examine how stimulation speed affects artificial afferent response, the data collection procedure is repeated twice at movements of 3\,mms\textsuperscript{-1} (slow) and 10\,mms\textsuperscript{-1} (fast). 

\subsubsection{Experiment 1b: Response to Bars, Edges and Gratings}

{\bf Stimuli:} A set of seven 3D-printed gratings G1-G7 (Figure \ref{robot_gratings}C and Table \ref{table1}) corresponding to those in the experimental study \cite{phillips2} are used. Bar and gap widths vary in the same proportions as those used in \cite{phillips2} with the scaling doubled. This scaling was chosen because the innervation density of markers in the TacTip ($\sim$70\,cm\textsuperscript{-2}) is around half that of SA-I afferents in the human fingertip. \\

\noindent {\bf Collection Procedure:} The TacTip is pressed successively onto a stimulus to a skin indentation of 1\,mm, with each press taking about 1\,sec and the sensor held stationary at the bottom of a press for 0.5\,sec. Between each press, the sensor is moved 0.2\,mm perpendicular to the bar/gap axis, starting in free space on the left-hand side of the stimulus and moving 90\,mm over the entire stimulus until reaching free space again. Tactile image data are recorded for the entire downward phase and stationary period of each press, once for each of the seven stimuli. 

\subsubsection{Experiment 2: Discrimination of Grating Orientation}
\label{sec:exp2}

{\bf Stimuli:} Seven square-wave grating stimuli were produced by cutting grooves (depth 1.5\,mm) in plastic blocks with a CNC milling machine (example depicted in Figure \ref{robot_gratings}B). The periodicity of the gratings, 1, 1.5, 2, 2.5, 3, 4 and 5\,mm, correspond to those used in the analogous psychophysical grating-discrimination experiment \cite{johnson}. Additionally, we use a smooth eighth surface considered as a grating of periodicity 0\,mm. \\

\noindent {\bf Collection Procedure:} Data are collected by pressing the TacTip skin onto the grating stimuli so that a compression of 2.5\,mm is achieved, after which it is held stationary for 1\,s. For training, the yaw angle, $\psi$, of the sensor relative to each grating is treated as a label, and randomly sampled from a uniform distribution $-90\degree\leq\psi\leq 90\degree$ by rotating the wrist joint of the robot. The extremes $\psi = \pm90\degree$ are arranged so that grating grooves align with the rows/columns of TacTip markers.  Roll $\phi$, pitch $\theta$, $x$, $y$ and $z$ are randomly sampled within ranges $\pm$2\degree, $\pm$2\degree, $\pm$2.5\,mm, $\pm$2.5\,mm and $\pm$0.15\,mm respectively, but not retained as labels for the data. In addition, the physical orientation of the grating is systematically changed through (90\degree, 180\degree, 270\degree, 360\degree), with the rotation angle $\psi$ relative to this orientation. This variability is to avoid systematic dependence on these dimensions with rotation angle if the sensor is not perfectly level relative to the stimulus, which the neural network could otherwise use to distinguish grating orientation. The training/validation set consists of 1000 samples per grating, giving a total of $N=8000$ samples. 

For testing, 300 samples per grating are collected at each of $\psi=-45\degree$ and $\psi=+45\degree$. We denote these as conditions A and B, such that stimuli A and B are perpendicular and equidistant from the $\pm90$\degree bounds of the training data. As with the training data, $\phi$, $\theta$, $x$, $y$ and $z$ are all randomly sampled within their above ranges. 

\setlength{\tabcolsep}{6.7pt}
\begin{table}
	\captionof{table}{Ridge and gap widths of the aperiodic gratings in Experiment 1b, based upon an investigation of spatial encoding in biological touch \cite{phillips2}.}
	{\begin{tabularx}{\linewidth}{p{0.5cm}p{1.1cm}p{1.3cm}p{0.8cm}p{0.1cm}p{0.5cm}p{0.5cm}p{0.5cm} p{0.5cm}p{0.5cm}p{0.5cm}p{0.5cm}p{0.5cm}p{2cm}}
		\multicolumn{1}{c}{} & \multicolumn{3}{c}{\textbf{\footnotesize Ridge widths (mm)}} & & \multicolumn{8}{c}{\textbf{\footnotesize Gap widths (mm)}} & \\
		\\[-1.5em]
		\multicolumn{1}{c}{\textbf{\footnotesize Grating}}&\multicolumn{1}{c}{\textbf{\footnotesize left}} & \multicolumn{1}{c}{\textbf{\footnotesize internal}} & \multicolumn{1}{c}{\textbf{\footnotesize right}} & & \multicolumn{8}{c}{\textbf{\footnotesize left to central to right}} & \multicolumn{1}{c}{\textbf{\footnotesize Overall length (mm)}}\\ 
		\hline
		\multicolumn{1}{c}{\textbf{\footnotesize G1}}&\multicolumn{1}{c}{\footnotesize 6.0} & \multicolumn{1}{c}{\footnotesize 1.0} & \multicolumn{1}{c}{\footnotesize 6.0} & & \multicolumn{1}{c}{\footnotesize 1.0} & \multicolumn{1}{c}{\footnotesize 1.0} & \multicolumn{1}{c}{\footnotesize 1.5} & \multicolumn{1}{c}{\footnotesize 2.0} & \multicolumn{1}{c}{\footnotesize 3.0} & \multicolumn{1}{c}{\footnotesize 4.0} & \multicolumn{1}{c}{\footnotesize 6.0} & \multicolumn{1}{c}{\footnotesize 10.0} & \multicolumn{1}{c}{\footnotesize 47.5}\\
		\multicolumn{1}{c}{\textbf{\footnotesize G2}}&\multicolumn{1}{c}{\footnotesize 6.0} & \multicolumn{1}{c}{\footnotesize 1.5} & \multicolumn{1}{c}{\footnotesize 6.0} & & \multicolumn{1}{c}{\footnotesize 1.0} & \multicolumn{1}{c}{\footnotesize 1.0} & \multicolumn{1}{c}{\footnotesize 1.5} & \multicolumn{1}{c}{\footnotesize 2.0} & \multicolumn{1}{c}{\footnotesize 3.0} & \multicolumn{1}{c}{\footnotesize 4.0} & \multicolumn{1}{c}{\footnotesize 6.0} & \multicolumn{1}{c}{\footnotesize 10.0} & \multicolumn{1}{c}{\footnotesize 51.0}\\
		\multicolumn{1}{c}{\textbf{\footnotesize G3}}&\multicolumn{1}{c}{\footnotesize 3.0} & \multicolumn{1}{c}{\footnotesize 2.0} & \multicolumn{1}{c}{\footnotesize 6.0} & & \multicolumn{1}{c}{\footnotesize 1.0} & \multicolumn{1}{c}{\footnotesize 1.0} & \multicolumn{1}{c}{\footnotesize 1.5} & \multicolumn{1}{c}{\footnotesize 2.0} & \multicolumn{1}{c}{\footnotesize 3.0} & \multicolumn{1}{c}{\footnotesize 4.0} & \multicolumn{1}{c}{\footnotesize 6.0} & \multicolumn{1}{c}{\footnotesize 10.0} & \multicolumn{1}{c}{\footnotesize 51.5}\\
		\multicolumn{1}{c}{\textbf{\footnotesize G4}}&\multicolumn{1}{c}{\footnotesize 6.0} & \multicolumn{1}{c}{\footnotesize 3.0} & \multicolumn{1}{c}{\footnotesize 6.0} & & \multicolumn{1}{c}{\footnotesize 1.0} & \multicolumn{1}{c}{\footnotesize 1.0} & \multicolumn{1}{c}{\footnotesize 1.5} & \multicolumn{1}{c}{\footnotesize 2.0} & \multicolumn{1}{c}{\footnotesize 3.0} & \multicolumn{1}{c}{\footnotesize 4.0} & \multicolumn{1}{c}{\footnotesize 6.0} & \multicolumn{1}{c}{\footnotesize} & \multicolumn{1}{c}{\footnotesize 48.5}\\
		\multicolumn{1}{c}{\textbf{\footnotesize G5}} &\multicolumn{1}{c}{\footnotesize 6.0} & \multicolumn{1}{c}{\footnotesize 4.0} & \multicolumn{1}{c}{\footnotesize 6.0} & & \multicolumn{1}{c}{\footnotesize} & \multicolumn{1}{c}{\footnotesize 1.0} & \multicolumn{1}{c}{\footnotesize 1.5} & \multicolumn{1}{c}{\footnotesize 2.0} & \multicolumn{1}{c}{\footnotesize 3.0} & \multicolumn{1}{c}{\footnotesize 4.0} & \multicolumn{1}{c}{\footnotesize 6.0} & \multicolumn{1}{c}{\footnotesize} & \multicolumn{1}{c}{\footnotesize 49.5}\\
		\multicolumn{1}{c}{\textbf{\footnotesize G6}} &\multicolumn{1}{c}{\footnotesize 6.0} & \multicolumn{1}{c}{\footnotesize 6.0} & \multicolumn{1}{c}{\footnotesize 6.0} & & \multicolumn{1}{c}{\footnotesize} & \multicolumn{1}{c}{\footnotesize} & \multicolumn{1}{c}{\footnotesize} & \multicolumn{1}{c}{\footnotesize 2.0} & \multicolumn{1}{c}{\footnotesize 3.0} & \multicolumn{1}{c}{\footnotesize 4.0} & \multicolumn{1}{c}{\footnotesize 6.0} & \multicolumn{1}{c}{\footnotesize} & \multicolumn{1}{c}{\footnotesize 45.0}\\
		\multicolumn{1}{c}{\textbf{\footnotesize G7}} &\multicolumn{1}{c}{\footnotesize 10.0} & \multicolumn{1}{c}{\footnotesize 10.0} & \multicolumn{1}{c}{\footnotesize 10.0} & & \multicolumn{1}{c}{\footnotesize} & \multicolumn{1}{c}{\footnotesize} & \multicolumn{1}{c}{\footnotesize} & \multicolumn{1}{c}{\footnotesize} & \multicolumn{1}{c}{\footnotesize} & \multicolumn{1}{c}{\footnotesize} & \multicolumn{1}{c}{\footnotesize 6.0} & \multicolumn{1}{c}{\footnotesize 10.0} & \multicolumn{1}{c}{\footnotesize 46.0}\\
	\end{tabularx}}
	\par\medskip
	\label{table1}
\end{table} 

\subsection{Analysis}

\subsubsection{Sample Preparation}

For training, validation and testing, artificial SA-I and RA-I responses are extracted off-line for each video frame using the methods in Section 2.3. SA-I and RA-I samples are tactile images of their respective afferent responses over the surface of the tip (Figure \ref{tactile_images_presses}). In both cases, only the frame in each press with the largest total response is used for analysis.

\subsubsection{Model of Grating-Orientation Perception}

For predicting the grating angle in Experiment 2, two convolutional neural network (CNN) regression models, CNN-SA-I and CNN-RA-I, are implemented for artificial SA-I and RA-I afferents. Both networks are trained to predict grating angle $\psi$ using their afferent responses extracted from training data on all seven gratings ($N_{\rm train}=5250$) and likewise validated on all seven gratings ($N_{\rm val}=1750$). \\

\begin{figure}[t!]
\centering
\includegraphics[width=\linewidth]{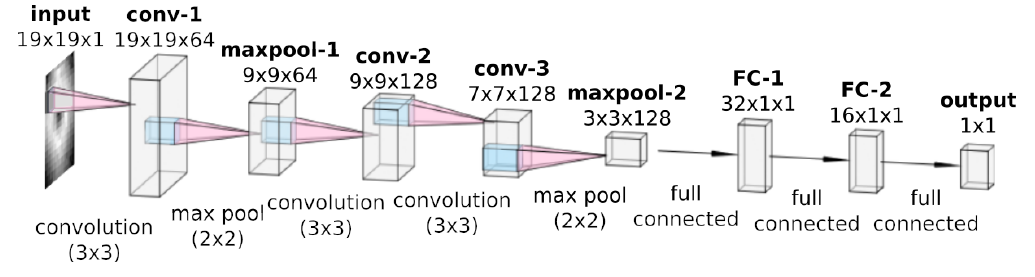}
\caption{CNN-SA-I and CNN-RA-I model architecture for artificial SA-I and RA-I afferents in the grating-orientation discrimination task (Experiment 2). } 
\label{cnn}
\end{figure}

\noindent{\bf Model Architecture:} The CNN-SA-I and CNN-RA-I models are constructed with a 2D convolutional neural network, built using Keras with TensorFlow backend (Figure \ref{cnn}). The same network architecture is used for both artificial afferent types using 19$\times$19 tactile images as input. A spatial feature extractor comprises an initial convolutional layer containing 64 3$\times$3 kernels; resulting feature maps (19$\times$19) are then down-sampled (9$\times$9) with a 2$\times$2 max-pooling layer; this is followed by two stacked convolutional layers each with 128 3$\times$3 kernels and a final 2$\times$2 max-pooling layer providing 128 3$\times$3 feature maps. The output is flattened and passed through two hidden dense layers with 32 and 16 neurons each and finally the output layer provides a continuous output via a single neuron with no activation function. All hidden layers use ReLU activation functions. Regularisation includes drop-outs of 0.4, 0.2 and 0.2 prior to the two hidden dense layers and the output layer respectively and L2-regularisation (factor 0.005) on each dense layer. Batch normalisation is implemented after each convolutional layer. Model hyperparameters and architecture were hand-tuned on the validation set. Training uses a batch size of 32 samples over about 50 epochs using a patience of 20 epochs having non-decreasing loss, of which the models with highest validation accuracy are retained for testing.

\subsubsection{Model of Two-Interval Same-Different Task}
\label{sec04:06:05}

A two-interval same-different task is a psychophysical paradigm where participants are presented with two intervals per trial, with each interval presenting either stimulus\,A or stimulus\,B, which here correspond to grating orientations of $\psi = \mp45\degree$ respectively (Section~\ref{sec:exp2}, Experiment 2). Participants respond by stating whether the stimuli in the two intervals were the same (denoted ${\rm R}^{(1)}$) or different (denoted ${\rm R}^{(2)}$) \cite{macmillan}. This setup mirrors a psychophysical grating experiment (Johnson and Phillips (1981) \cite{johnson}), where the possible presentations on each trial are either AA or AB, called ${\rm S}^{(1)}$ and ${\rm S}^{(2)}$ respectively. 
Overall, the experiment probes the capability of the subjects to perceive grating orientation dependant upon the grating period.

Signal detection theory (SDT) is used to relate perception to sensitivity of sensory-driven neural activity. Central to SDT is the decision variable, which is an interpretation of neural activity that guides decision making \cite{gold}. The process of forming a decision variable from neural activity is often referred to as decoding (represented in Figures \ref{bioT}A, \ref{bioT}B by the decoding phase), with the decision variable then compared to a criterion value to make a decision. Our approach to robot decision making for the two-interval same-different task equates the continuous output of the trained SA-I and RA-I regression models of the grating orientation $\psi$ to the decision variable $s$ in SDT. Training labels were normalised to fall between 0 and 1 prior to training; therefore, stimuli A and B correspond to $s=0.25$ and $0.75$ on the scale of the decision variable. In practice, probability distributions, $f(s{\mid}{\rm A})$ and $f(s{\mid}{\rm B})$, are constructed from the model outputs  (Figure \ref{plots}A\textsubscript{1-3}), from which individual two-choice covert decisions for each interval are found \cite{macmillan,decarlo}. 

In the context of the presented task, the robot decides if stimulus\,A ($\psi=-45\degree$) or stimulus\,B ($\psi=+45\degree$) was presented in interval one and subsequently again for interval two. The overall decision of ``same'', ${\rm R}^{(1)}$, or ``different'', ${\rm R}^{(2)}$, is based on relative covert decisions for the two intervals. The first step, therefore, is to perform the two-choice task with stimuli A and B as the two mutually-exclusive alternatives.
These covert decisions then give the true-positive $\rm{P}(\rm{R}^{(2)}{\mid}\rm{S}^{(2)})$ and false-positive $\rm{P}(\rm{R}^{(2)}{\mid}\rm{S}^{(1)})$ response rates for the two-interval same-different task, where ``different" is regarded as a positive response. \\

\noindent{\bf Decision rule:} The probability of the two covert decision alternatives, ``a'' or ``b'', conditional on the presented stimulus, A or B, can be determined from the area under the probability distribution of the decision variable, $s$, for that stimulus, either side of the criterion value, \textit{c} (Figure \ref{plots}A\textsubscript{1-3}). Probability distributions, $f(s{\mid}{\rm{A}})$ and $f(s{\mid}{\rm{B}})$ (Figure \ref{plots}A\textsubscript{1-3}), are constructed from the output of the trained SA-I and RA-I regression model when presented with all 600 test samples at $\psi = \mp45\degree$. A simple decision rule is used to maximise the percentage of correct responses, \textit{i.e.} to equally reward correct responses (a given A and b given B) and equally penalise incorrect responses (a given B and b given A), which was to respond ``a'' if the likelihood ratio ${\rm LR}(s)=f(s{\mid}{\rm A})/f(s{\mid}{\rm B})>1$ and to respond ``b'' if ${\rm LR}(s)<1$ \cite{green}. This corresponds to a criterion value, $c$, which is shown in Figures \ref{plots}A\textsubscript{1-3} by the dashed vertical line where the distributions intersect.

Given the conditional probabilities of each outcome in covert decisions (Figure \ref{plots}A\textsubscript{1-3}), the response probabilities are computed as follows:
\begin{eqnarray}
\rm{P}(\rm{R}^{(2)}{\mid}\rm{S}^{(2)}) = \rm{P}(a{\mid}\rm{A})\rm{P}(b{\mid}\rm{B}) + \rm{P}(b{\mid}\rm{A})\rm{P}(a{\mid}\rm{B}),
\label{p1}\\
\rm{P}(\rm{R}^{(2)}{\mid}\rm{S}^{(1)}) = \rm{P}(a{\mid}\rm{A})\rm{P}(b{\mid}\rm{A}) + \rm{P}(b{\mid}\rm{A})\rm{P}(a{\mid}\rm{A}),
\label{p2}
\end{eqnarray}
where $\rm{P}(\rm{R}^{(2)}{\mid}\rm{S}^{(2)})$ and $\rm{P}(\rm{R}^{(2)}{\mid}\rm{S}^{(1)})$ are computed for each grating period and used to create a robot-psychometric function (Figure \ref{plots}B).

\section{Results}
\label{sec02}

\begin{figure}[t]
\centering
\includegraphics[width=\linewidth]{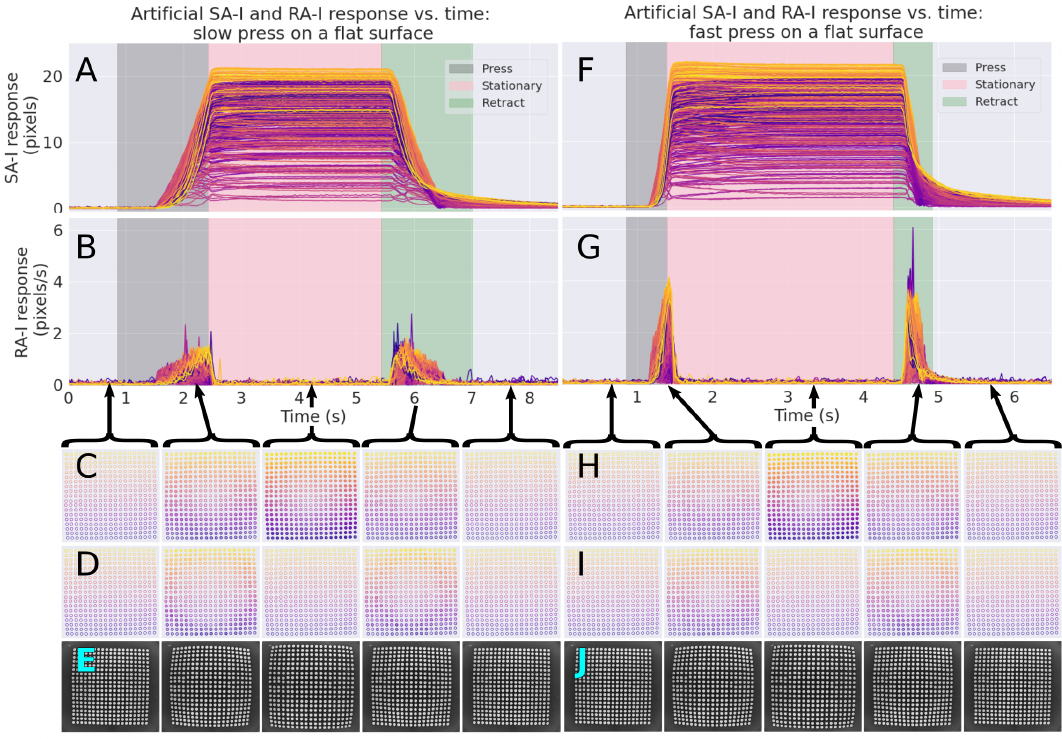}
\caption{Artificial SA-I, RA-I responses collected during a press on a flat surface at slow 3\,mms$\textsuperscript{-1}$ (panels \textbf{A}-\textbf{E}) and fast 10\,mms$\textsuperscript{-1}$ speeds (panels \textbf{F}-\textbf{J}). (\textbf{A}, \textbf{F}; \textbf{B}, \textbf{G}) show the SA-I and RA-I responses vs time for slow and fast presses respectively. (\textbf{C}, \textbf{H}; \textbf{D}, \textbf{I}) show the spatial response profiles of \mbox{SA-I} afferents and RA-I afferents at the indicated times; the opacity (fill) of the marker denotes activity coloured by its location in the array. (\textbf{E}, \textbf{J}) show the original camera frames for those responses.}\label{plate_slow}
\end{figure}

\subsection{Experiment 1a: Response to Normal Pressure}
\label{sec02:01}

This experiment was adopted to test whether the proposed artificial SA-I and RA-I afferent models resemble their natural counterparts when stimulated with a simple normal force applied by a flat surface. First, we consider adaption rates and compare the response of artificial type-I afferents to the firing rates of individual natural type-I afferents from several studies that use the same stimulation conditions \cite{saal3,jenmalm,mountcastle2,knibestol2}. Following this comparison, we consider the spatial modulation of artificial afferent responses across the entire population. We asses the nature of the skin dynamics to which individual artificial afferents respond and make an initial hypothesis about the capability of these artificial tactile channels to mediate population codes.     

\subsubsection{Artificial Type-I Afferents Model Natural Afferent Adaption Rates}
\label{sec02:01:01}

Artificial SA-I afferents responded to the stimulus onset and hold phase but not the offset (Figures \ref{plate_slow}A,F). Artificial RA-I afferents responded strongly to the stimulus onset and offset but not the hold phase (Figures \ref{plate_slow}B,G). These patterns of activity are strongly indicative of their natural counterparts, which we now examine in more detail.

Natural SA-I afferent firing rates are known to increase with indentation depth \cite{mountcastle2}. This correlation between indentation depth and firing rate is mirrored in the artificial SA-I afferents, which demonstrate a trapezoidal response curve with steady activity over the hold phase increasing with indentation depth (Figures \ref{plate_slow}A,F; both slow and fast presses). Further supporting the likeness of artificial SA-I afferents to their natural counterparts, both natural and artificial SA-I afferent responses fell to zero with removal of the stimulus \cite{jenmalm,mountcastle2}. 

Natural SA-I afferents are known to fire maximally during the dynamic phase of the stimulus presentation, with a slow reduction of activity under constant stimulation (slow adaption)~\cite{jenmalm}. The artificial SA-I afferents diverged from this: the artificial SA-I response rose steadily as the TacTip was pressed onto the stimulus, then had sustained maximal activity without slow adaption. We comment that all biological sensory systems adapt to stimulation for reasons such as their molecular signalling pathways; however, artificial sensors are designed for repeatable operation where `no adaption' rather than `slow adaptation' is desired.    

Natural RA-I afferents are known to respond only during dynamic phases of stimulus presentation, rapidly adapting during stimulus onset and offset \cite{jenmalm}. In accordance, artificial RA-I afferents also exhibited phasic responses: for both slow and fast presses, artificial RA-I channels responded only during the dynamic phases of contact (Figures \ref{plate_slow}B,G; slow and fast presses). Furthermore, the peak response of the artificial RA-I afferents was greater for fast presses akin to natural RA-I afferents, which show raised firing rates with increased indentation speed \cite{knibestol2}. 

\subsubsection{Individual Artificial Type-I Afferents Encode Local Shear and Population Response Encodes Shape}
\label{sec02:01:02}

Both the artificial SA-I and RA-I afferents are good candidates for providing viable codes to represent static extended stimuli via spatial modulation, because the spatial arrangement of artificial responses provides information about the contact shape akin to natural afferents \cite{srinivasan2,goodwin}.

\begin{figure}[t]
	\centering
	\includegraphics[width=0.8\linewidth]{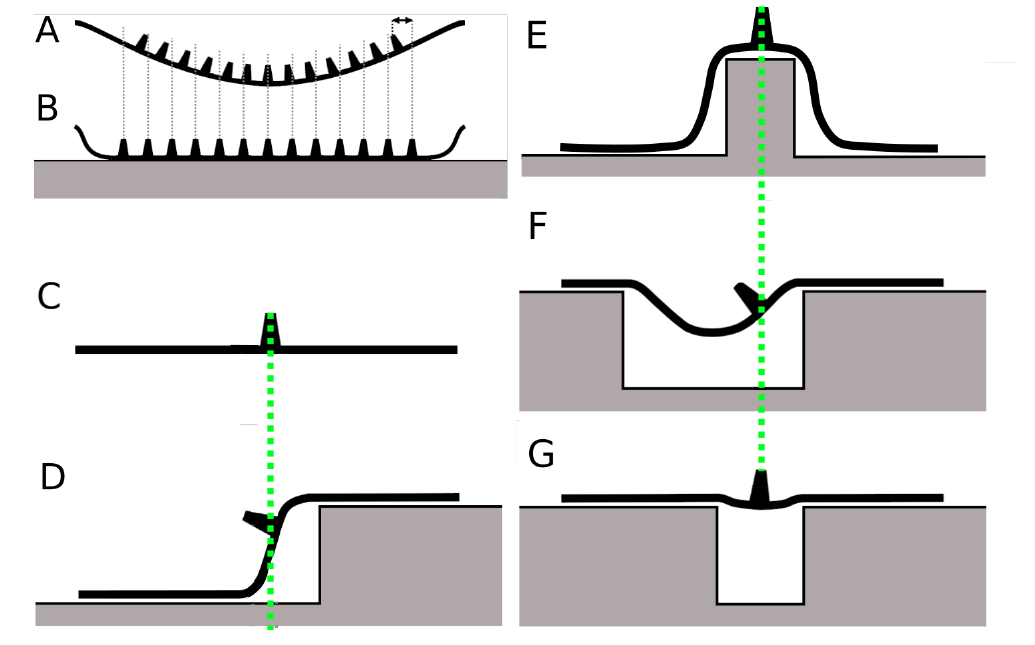}
	\caption{(\textbf{A},\textbf{B}) Physical model of TacTip skin deformation when pressed onto a flat~plate. (\textbf{A}) At rest, \textit{i.e.} no deformation, and (\textbf{B}) pressed flat. (\textbf{C}-\textbf{G}) Diagram of the TacTip's central artificial afferent pressed onto features of aperiodic gratings: at rest (\textbf{C}), an edge (\textbf{D}), an isolated bar (\textbf{E}), and wide (\textbf{F}) or narrow gaps (\textbf{G}) between two edges.}
	\label{models}
\end{figure} 

The spatial arrangement of the artificial SA-I population response shows an approximately circular central region where minimal activity occurs, even at times of peak stimulation (Figures \ref{plate_slow}C,H; slow and fast presses). Radiating away from this central region, afferents exhibit increased activity. This spatial pattern can be understood by considering a basic physical model of the interaction between the biomimetic fingertip’s skin and the stimulus: when the slightly-rounded tip is pressed onto a flat surface, a circular area of the tip compresses and conforms to the surface (Figures \ref{models}A,B). The size of this area depends on how strongly the fingertip is pressed onto the surface, with the markers away from the center spreading outwards as the skin flattens. Thus, the central artificial SA-I afferent has minimal activity and the off-centre afferents in a ring about that centre have raised activity.

At peak amplitudes of the artificial RA-I response, during the dynamic phase, the activity is also low in the central region and raised in a ring around that centre (Figures \ref{plate_slow}D, I; 2nd and 4th images), giving an approximately radial population response. According to our physical model (Figures \ref{models}A,B), each marker moves in one direction during the press phase (towards the circumference) and in the opposite direction during the release phase (towards the centre). The artificial RA-I afferent response is a measure of marker displacement per frame, which is the local shear-velocity magnitude. Thus, in this particular example, the spatial arrangements of artificial SA-I and RA-I responses are similar. Note though that their peak responses take place at different times: SA-I afferents respond most during the static phase of the press and RA-I afferents respond most in the middle of the dynamic phase.

\subsection{Experiment 1b: Response to Bars, Edges and Gratings}
\label{sec02:02}

This experiment mirrors a classic (1981) neurophysiological study into peripheral neural representations of spatially-complex tactile stimuli (edges and bars) in monkey primary afferents by Phillips and Johnson \cite{phillips2}. Their study was accompanied with a psychophysics study into human perception of tactile stimuli (including gratings) via spatial neural mechanisms \cite{johnson}, with both studies aiming to understand the ``coding mechanisms underlying the human’s ability to resolve gratings".

Here we compare the activity of centrally-located artificial type-I afferents with natural type-I afferent recordings taken from \cite{phillips2} under similar complex stimulation with grating stimuli (Table 1). In addition to comparing artificial and natural mechanoreception, we use this experiment to understand our biomimetic tactile sensor's ability to spatially resolve grating stimuli in accordance with the neurophysiology of touch. Results are presented for three stimulus features: {(i)} sensitivity to edges; {(ii)} sensitivity to bars; and~{(iii)} the effect of neighbouring edges and bars. Results are displayed as spatial-response profiles (SRPs) for a single afferent, comprising the peak response over each press aggregated over many positions across the stimulus to give a spatial profile (Figure \ref{sa-i_grating}). 

\subsubsection{Sensitivity to Edges}
\label{sec02:02:01}

\begin{figure}[t]
    \vspace{-1em}
	\centering
	\includegraphics[width=\linewidth]{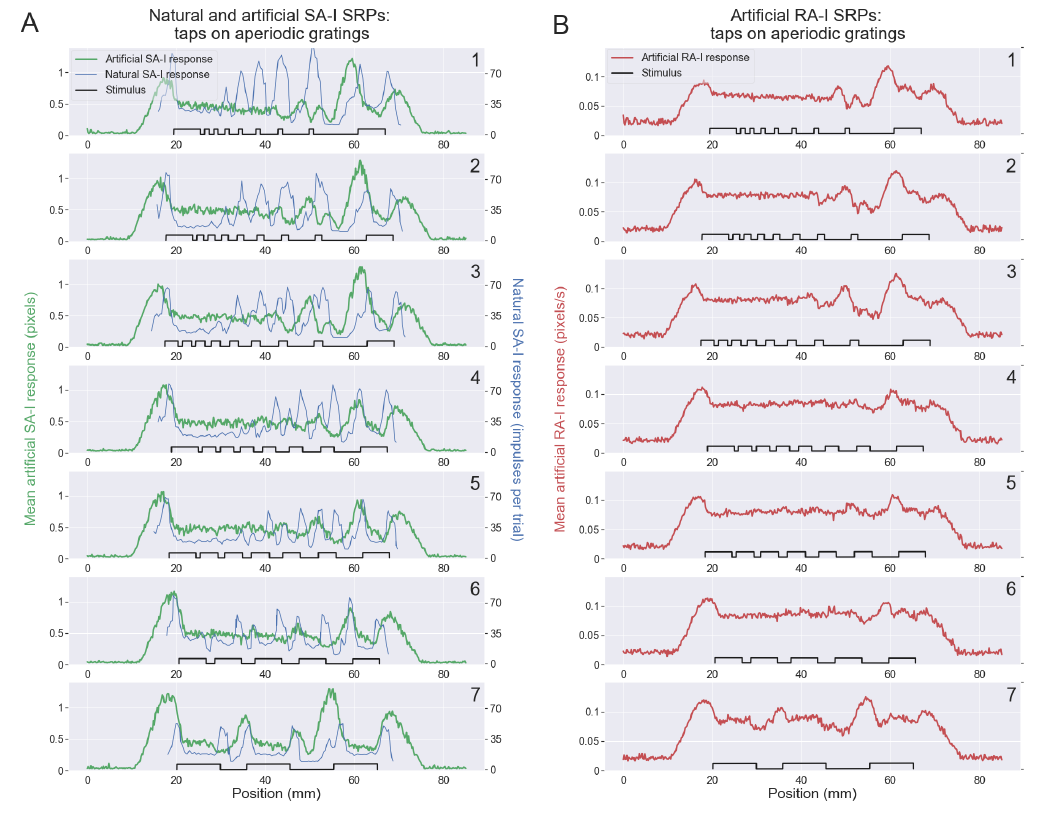}
	\vspace{-0.5em}
	\caption{Left: Spatial-response profiles artificial SA-I afferent (green) collected on gratings (black) compared with a single monkey SA-I afferent collected on seven gratings taken from Phillips and Johnson \cite{phillips2}. Right: SRPs for the central RA-I afferent (red) collected on gratings (black). The scale for natural afferents is doubled compared with \cite{phillips2}. Note that these responses are from distinct phases of the press (left, static; right, dynamic, respectively). As expected, human touch is more sensitive than artificial touch, but otherwise the SRPs are similar on bars, edges and gaps.}
	\label{sa-i_grating}
\end{figure}

The central artificial SA-I afferent exhibits preferential sensitivity to ``edges facing a large gap" (Figure \ref{sa-i_grating}A), which is a characteristic of natural SA-I afferents \cite{phillips2}. This pattern is most clearly seen in panel A7, where the response of natural and artificial SA-I afferents were attenuated by three bars and amplified by the six associated edges. This also occurs on the approach and departure for each of the other gratings, where the gratings exhibit rising or falling edges before and after free space. 

The central artificial RA-I afferent has markedly less sensitivity to edges than artificial SA-I afferents (Figure~\ref{sa-i_grating}B). (Note that SRPs are not shown for natural RA-I afferents because they were only available in \cite{phillips2} for some stimuli.) This seems to be in accordance with artificial RA-I afferents, which did not show any edge enhancement in those experiments \cite{phillips2}. That said, there may be some sensitivity to edges at the extremities of the stimulus, but there are subtleties in this comparison because the displayed artificial SA-I and RA-I activities are at distinct phases of the press: the artificial SA-I activity is during the static phase when the pressure is maximal, and the artificial RA-I activity during the initial dynamic phase when the rate of change of pressure is maximal.

In the case of artificial afferents, edge enhancement can be explained with our physical model of local shear strain within the biomimetic fingertip (Figure \ref{models}). As the skin bends around an edge, the marker undergoes shear strain due to levering of the internal pins, which causes the artificial SA-I afferents to respond (Figures \ref{models}C,D).

Phillips and Johnson hypothesised that the difference in SRPs between natural SA-I and RA-I afferents was a consequence of an ``intrinsic difference in the spatial organization of their receptor mechanisms'' \cite{phillips2}. However, in our study the spatial arrangement of receptive fields of artificial SA-I and RA-I afferents are necessarily identical, being derived from the same markers. \textcolor{black}{Why then do the artificial RA-I afferents still respond less strongly?} \textcolor{black}{Instead, we attribute this effect in the artificial afferents to small perturbations in marker position during the initial and final stages of the press. The preprocessing of RA-I afferent activity amplifies those perturbations owing to the derivative (Equation~2), giving less edge enhancement due to a higher noise floor (also evident in Experiment 2 below).}

\subsubsection{Sensitivity to Bars}
\label{sec02:02:02}

A consequence of amplified edge response, seen in both natural SA-I and our artificial afferents, is a comparatively diminished response to isolated bars. 
The exclusive response of artificial type-I afferents to edges can be understood with the physical model of the biomimetic fingertip described above (Figure \ref{models}). Locally, bars are considered flat surfaces and thus the artificial afferent should not produce any response when located directly above the bar (Figure \ref{models}E). In practice, however, some response is still observed, which for artificial RA-I afferent we attribute to the noise floor described above and for artificial SA-I afferents to shear during the contact.

\subsubsection{Effect of Neighbouring Bars and Edges}
\label{sec02:02:02}

The sensitivity of both individual natural and artificial type-I afferents to edge and bar features was attenuated with reduced gap width. Both natural and artificial SA-I afferents show reduced edge amplification from right to left on the aperiodic grating stimuli G1-G7 as gaps between the edges become smaller (Figure \ref{sa-i_grating}A). 

For artificial afferents, this attenuation can be explained by extending our physical model of the biomimetic fingertip (Figure \ref{models}). The amount of marker deflection is reduced with decreasing gap width owing to limited flexibility of the artificial skin. In essence, the skin acts as a low-pass filter that attenuates its response to high spatial frequencies of the stimulus (Figures~\ref{models}E, F, G).  

In the original physiological experiments, it was observed that ``as the bars are spaced more closely than 3.0\,mm the heights of the associated response peaks are diminished"  \cite{phillips2}. Likewise, the artificial SA-I afferents show diminished response peaks for narrower gap widths~(Figure \ref{sa-i_grating}A). This effect is more pronounced for the artificial afferents than their natural counterparts because of a larger length scale where artificial edge responses begin to be diminished (visible at 10\,mm in aperiodic gratings G1, G2, G3 and G7, and 6\,mm in aperiodic gratings G4, G5 and G6). 

Both natural and artificial RA-I afferents also exhibit attenuation in the stimulus detail represented within their associated SRPs with reducing gap width (Figure \ref{sa-i_grating}B for artificial RA-I afferent SRPs; see \cite{phillips2} for natural RA-I afferent SRPs). Furthermore, both natural and artificial RA-I afferents exhibit faster attenuation in response to reducing gap width than their SA-I counterparts. For natural RA-I afferents, ``the profiles of all five fibers were clearly modulated by the 5.0\,mm gap in the stimulus, but gaps of 0.5-1.0\,mm were not represented in any of the five profiles" \cite{phillips2}. The central artificial RA-I afferent profile similarly shows modulation to 10 and 6\,mm gaps but edges bounded by gaps of 1-4\,mm are not represented due to the lower sensitivity of the tactile sensor.

\subsection{Experiment 2: Discrimination of Grating Orientation}
\label{sec02:03}

\begin{figure}[t]
    \begin{center}
		\includegraphics[width=\linewidth]{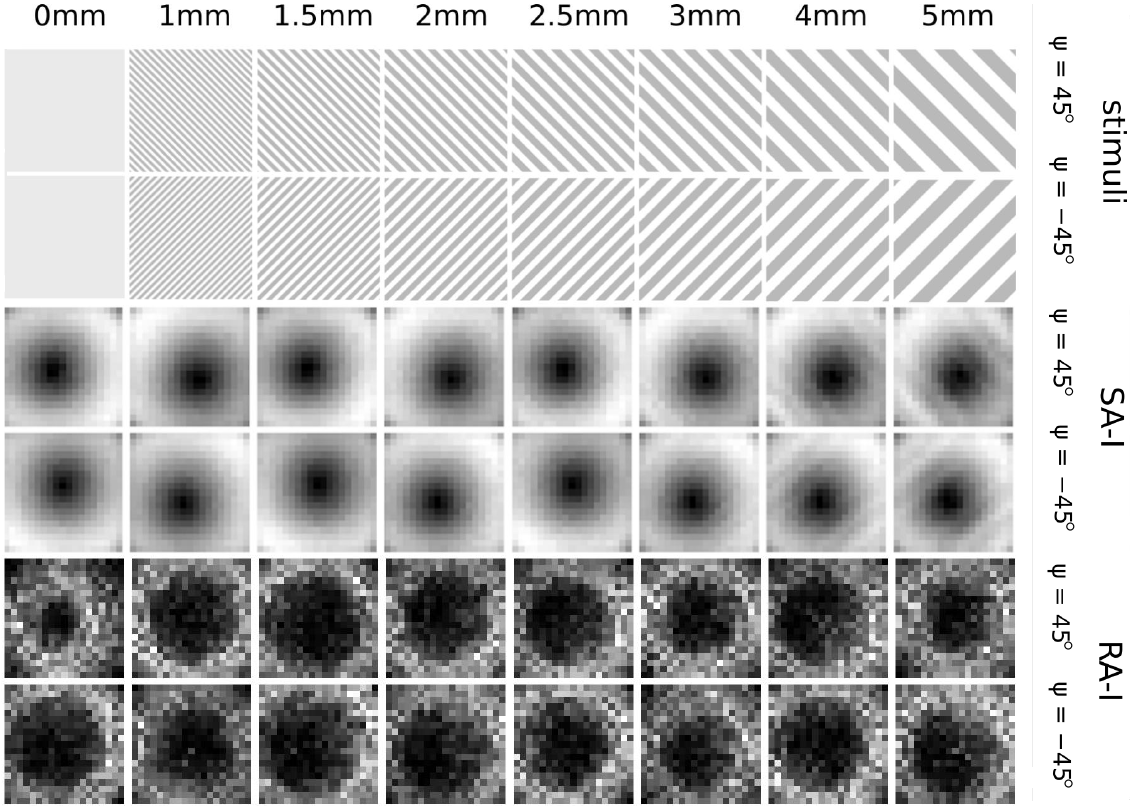}
        \caption{Examples of artificial SA-I and RA-I afferent population responses (centre and bottom rows) collected on grating stimuli (same scale; top rows). For each grating period (0-5\,mm), both possible orientations ($\pm\ang{45}$) are displayed (upper/lower). The colour of each pixel in the tactile image represents the response magnitude of an artificial afferent: lighter corresponds to larger responses.}
		\label{tactile_images_presses}
		\end{center}
\end{figure}

This experiment mirrors a classic (1981) two-interval same-different psychophysical task by Johnson and Phillips \cite{johnson}, where the ability for human participants to discriminate grating orientation is affected by the spatial period of the grating stimuli. Here we use the same set of seven periodic grating stimuli with grating periods 1-5\,mm and an additional eighth stimulus with grating period 0\,mm (smooth) as a control (Figure \ref{tactile_images_presses}; left column). Grating orientation was of particular interest because non-spatial stimulus features, such as total stimulus area and edge content, are consistent between the first and second interval of each trial, which controls the use of non-spatial mechanisms for the discrimination \cite{johnson}. For this reason, many experimenters have subsequently used grating orientation as a tool for measuring tactile spatial acuity \cite{vanboven,vanboven2,tong}.  

\subsubsection{Artificial SA-I Spatial Structure Increases with Grating Period}

First, we consider the structure of artificial SA-I tactile images stimulated by gratings, with regard to their capability to spatially-resolve grating stimuli. Apparently, the spatial-encoding of the artificial SA-I afferent population is attenuated by reducing the grating period: there is visible fine spatial structure only for the coarser gratings with periods of 3\,mm or more (Figure \ref{tactile_images_presses}, middle rows). SA-I tactile images with periods up to 2.5\,mm exhibit a diminished structure: all have a central dark region of low response within an active surround and no obvious visible features indicating grating orientation. The only clear differences are the (unimportant) locations of the central region due to the experimental protocol of perturbing randomly the contact location and angle.

This attenuation in spatial structure of artificial SA-I tactile images appears consistent with the same effect that produced spatial attenuation to bars and edges in Experiment 1b above (Section~\ref{sec02:02:02}). In Experiment 1b, we explained the attenuation with a physical model of the biomimetic fingertip where the limited flexibility of the skin prevents it from conforming to narrow ridges (Figure \ref{models}). 

The grating period at which attenuation occurs in Experiment 2 (2.5\,mm period; 1.25\,mm spacing) is finer than in Experiment 1b (6\,mm period; 3\,mm spacing in aperiodic gratings G4, G5 and G6). However, there is a key difference between the two experiments. In Experiment 2, the most salient regions of the tactile images are the top-right/bottom-left (grating orientation $\psi=-\ang{45}$) and top-left/bottom-right (grating orientation $\psi=+\ang{45}$), furthest from the centre of the tactile image where the grating is oriented axially. Meanwhile, the grating in Experiment 1b is oriented top-to-bottom, leaving a small sensitive region at the left and right of the tactile image, with the sampled afferent taken from the center.

\begin{figure}[t]
\begin{center}
	\includegraphics[width=\linewidth]{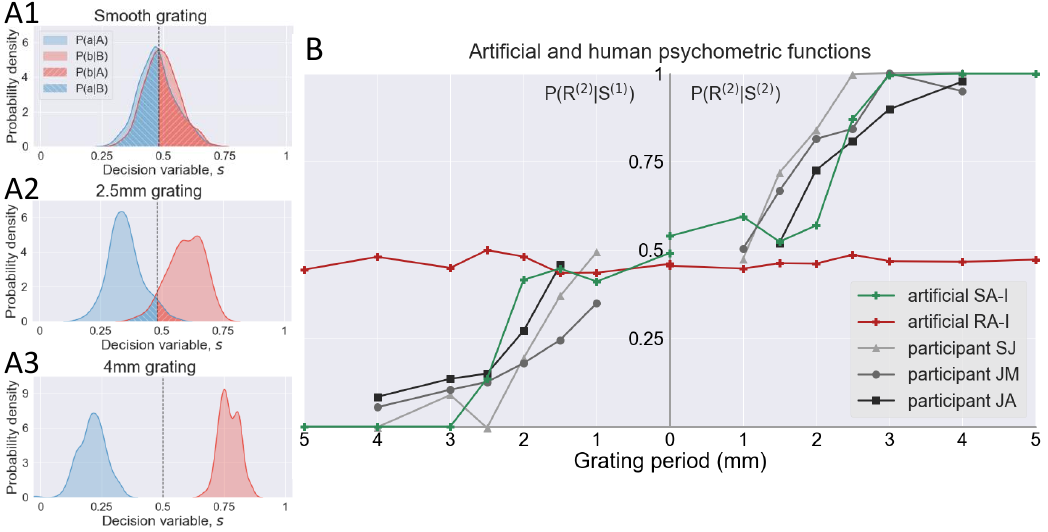}
	\caption{(\textbf{A1}, \textbf{A2}, \textbf{A3}) Covert decision models for artificial SA-I afferents in the grating orientation two-interval same-different task, with grating periods of 0\,mm, 2.5\,mm and 4\,mm respectively. Decisions ``a", ``b" correspond to $\psi=-45^\circ$ and $+45^\circ$ according to whether the decision variable $s$ is below or above the criterion value, $c$ (see methods; note that the light blue/red shaded regions of correct decisions extend to the criterion value). (\textbf{B}) Artificial SA-I/RA-I and human psychometric functions with grating period for the two-interval same-different task, according to the conditional probabilities for responding ``different", $\rm{R}^{(2)}$, given a different stimulus presentation in the second interval, $\rm{S}^{(2)}$ (right plots), or given the same stimulus presentation, $\rm{S}^{(1)}$ (left plots).} 
	\vspace{-0em}
	\label{plots}
\end{center}
\end{figure}


\subsubsection{Artificial RA-I Images do not Resemble Grating Orientation}
\label{sec02:03:02}

Unlike artificial SA-I images, the RA-I images do not exhibit any visible spatial structure indicative of grating orientation (Figure \ref{tactile_images_presses}, bottom rows). All artificial RA-I tactile images appear indistinguishable (aside from unimportant variations in the location and size of the central dark region discussed above), and are far noisier than the artificial SA-I images which would obscure any fine spatial detail. 

This lack of {grating-like} structure in artificial RA-I tactile images seems related to the relative lack of spatial detail exhibited in single artificial RA-I afferent activity in Experiment 1b (Figure \ref{sa-i_grating}B) compared to artificial SA-I activity (Figure \ref{sa-i_grating}A). Note also that tactile responses are shown when they each have their largest total responses. The peak response for artificial RA-I afferents is during the earlier dynamic phase and for artificial SA-I afferents during the static phase at maximal press. Thus, the lower pressure when RA-I activity peaks may contribute to its lower spatial sensitivity.

\subsubsection{Artificial and Human Performance on a Same-Different Grating-Orientation Task}

As described in the methods (Section~\ref{sec:exp2}), we followed Ref.~\cite{johnson} in implementing a robot psychophysical version of a two-interval same-different task that judges ``whether two gratings with the same period, presented sequentially, were presented with the same alignment, $\rm{S}^{(1)}$, or with different (orthogonal) alignments, $\rm{S}^{(2)}$''.  Our approach to robot decision making uses artificial CNN spatial decoders to estimate grating orientation $\psi$ from the SA-I and RA-I tactile images, comprising individual covert decisions that are combined sequentially for the overall same-different task. 

Distributions of covert decisions, a and b, for whether the grating orientation is $\psi=-45^\circ$ (A) or $+45^\circ$ (B) are each unimodal and intersect midway along the decision variable range, as required for an appropriate model (Figures \ref{plots}A$_1$-A$_3$; SA-I afferents only). As the grating period increases from 0-5\,mm, the discrimination becomes easier, as is apparent in the decreasing sizes of the incorrect choices,  ${\rm P(a{\mid}B)}$ and ${\rm P(b{\mid}A)}$, and increasing size of the correct choices, ${\rm P(a{\mid}A)}$ and ${\rm P(b{\mid}B)}$. The $0$\,mm (smooth) grating period serves as a control to check that features unrelated to grating orientation are not used for discrimination, which is confirmed by the similar size of the correct/incorrect regions.

For the two-interval same-different task, the conditional probabilities $\rm{P}(\rm{R}^{(2)}{\mid}\rm{S}^{(2)})$ for responding ``different'', ${\rm R}\textsuperscript{(2)}$, on differing stimuli, $\rm{S}^{(2)}$, increase as the grating becomes coarser and therefore more discriminable (Figure \ref{plots}B, right; SA-I/RA-I afferents in green/red). Likewise, the conditional probabilities $\rm{P}(\rm{R}^{(2)}{\mid}\rm{S}^{(1)})$ for responding ``different'', $\rm{R}^{(2)}$, on the same stimuli, $\rm{S}^{(1)}$, decrease for coarser, more discriminable gratings (Figure \ref{plots}B, left). 

For artificial SA-I afferents, there is a pronounced increase/decrease in the conditional probabilities, $\rm{P}(\rm{R}^{(2)}{\mid}\rm{S}^{(2)})$ and $\rm{P}(\rm{R}^{(1)}{\mid}\rm{S}^{(2)})$, from near 0.5 for completely smooth gratings (period 0\,mm) to near 0~($\rm{S}^{(2)}$) or 1~($\rm{S}^{(1)}$) on the coarser gratings of period 3\,mm or more. In contrast, for artificial RA-I afferents, there is no improvement over the central baseline level.

Overall, the sigmoid shape of the artificial SA-I curves in Figure \ref{plots}B is similar to human psychometric data, as shown here by superimposing experimental results from 3 human participants on the same task (grey curves, data from Ref.~\cite[Fig. 5]{johnson}). The poor discrimination from RA-I compared to SA-I artificial afferents is consistent with the experimental literature on human touch. For example, using vibrating grating stimuli to enhance the response of RA-I afferents resulted in little performance improvement \cite{johnson}, consistent with a minor role in this task.

In psychophysics, the just-noticeable difference (JND) is a standard metric of perceptual performance, corresponding to the stimulus difference on discrimination tasks that results in 75\%-correct decision rates on sigmoid-shaped psychometric curves. In the grating orientation task, the JND is 2.4\,mm for artificial SA-I afferents, which is comparable to 3 human participants in the range 1-7-2.1\,mm (Figure \ref{plots}B, grey curves). In addition, the artificial SA-I afferents gave near-perfect performance for grating periods $\geq$3\,mm, compared with performance of the three human participants who reached near-perfect performance at 2.5-4\,mm. Overall, human performance is somewhat better on the finer gratings, as expected from Experiment 1b, but otherwise the artificial and human behaviour are qualitatively similar on this task.

\section{Discussion}
\label{sec03}

This study explicitly considered the spatial encoding of artificial tactile type-I afferents to help understand their role in human and robot touch. Using the TacTip soft biomimetic tactile sensor with a skin based on the physiological structure of internal ridges and dermal papillae upon which markers are placed, we modelled the individual artificial SA-I and RA-I afferents as derived from marker displacement and marker speed. These artificial encodings mimicked the behaviour of natural type-I afferents in their adaption rates, spatial modulation and response to spatially-complex ridged stimuli. 

\subsection{Experiment 1a: Response to Normal Pressure}

Our first experiment considered individual artificial SA-I and RA-I responses to pressing on a flat plate at fast and slow speeds (Figures~4A, B, F, G). Multiple aspects of natural type-I afferent responses were mirrored by their artificial counterparts. Both artificial and natural SA-I afferent activity increased with indentation depth then fell to zero with removal of the stimulus to form characteristic trapezoidal response curves. Both artificial and natural RA-I activity responded only during the dynamic phases of the pressing motion, with peak responses greater for faster presses. A notable difference, however, is that natural SA-I afferents slowly reduce their activity during constant stimulation, but our model of artificial SA-I afferents displayed no adaption. While it would be fairly easy to capture slow adaption within a more complex SA-I model, we sought instead to provide the simplest possible model that captures the salient aspects of natural activity. Moreover, all biological sensory systems adapt to stimulation for reasons such as their use of molecular signaling pathways, whereas artificial sensors are designed for repeatable operation where slow adaptation may be undesirable. 

We then examined how the population response of artificial type-I afferents represents contact over individual afferents that each encode local shear. When pressing against a flat plate, the spatial arrangement of the peak SA-I and RA-I population responses showed an off-centre, on-surround pattern of activity (Figures~4C, D, H, I; also Figure 7). This spatial pattern is consistent with a physical model of the interaction between the slightly-rounded fingertip surface and the local shear-sensing of the artificial afferents: the centre of the fingerpad does not shear during normal indentation, while the surrounding circular region undergoes local shearing as the skin bends and levers the afferents (Figure~5). For artificial SA-I afferents, this spatial pattern peaks during the static phase of the contact at maximum pressure, and for RA-I afferents during the dynamic phase with changing pressure.

Most artificial tactile sensors transduce surface pressure into signals (e.g. via capacitive arrays) rather than shear~\cite{dahiya_tactile_2010}. However, biological SA-I and RA-I afferent responses are known to correlate with compressive and tensile strain~\cite{phillips3} with mechanoreceptors responsive to stretch and rate of stretch of their membranes~\cite{sripati}. Also, shear strain in skin may implicitly perform haptic edge detection by mapping gradients of the stimulus, identified by zero crossings of the shear-strain profile \cite{platkiewicz}. It is not obvious to us how a single mechanoreceptor can signal both the magnitude and direction of a physical quantity, as there is still debate on mechanisms for orientation selectivity in natural touch (unlike natural vision where orientation coding is well understood). For simplicity, therefore, we modelled artificial afferents using marker displacement magnitude only, but it would be straightforward to use directional components of shear strain as in other work with the TacTip~\cite{lepora2021}.

\subsection{Experiment 1b: Response to Bars, Edges and Gratings}

Our next experiment mirrored a classic (1981) study into peripheral neural representations of spatially-complex tactile stimuli (aperiodic edges and bars) in monkey primary afferents~\cite{phillips2}. We compared the activity of centrally-located artificial type-I afferents with natural electrophysiological recordings under similar complex stimulation over aperiodic grating stimuli. Both artificial and natural afferent activity exhibited preferential sensitivity to ``edges facing a large gap'' (Figure 6A). Both artificial and natural afferents also displayed a comparatively diminished response to isolated bars as a consequence of the amplified edge response. Furthermore, the sensitivity of both artificial and natural afferents to edge and bar features was attenuated with reduced gap width. Artificial RA-I afferents had reduced sensitivity attributed to a higher noise floor from the derivative in the RA-I model (Equation~2), while measured artificial RA-I activity exhibits no edge enhancement~\cite{phillips2}. 

Given that our model of the biomimetic fingertip's skin (Figure \ref{models}) implies that individual artificial SA-I and RA-I afferents encode local shear strain, the spatial modulation of each artificial afferent response will depend on the local stimulation. The effect of neighbouring edges and bars appears similar for natural and artificial type-I afferents, except that the length scale for artificial afferents is larger (6-10\,mm) than for natural afferents ($\sim\!3$\,mm). The lower sensitivity of the artificial afferents appears due to the artificial skin being thicker and less flexible than its natural counterpart with a lower artificial afferent density. Other than this difference in sensitivity, there was a high degree of convergence between the artifical and natural systems.

\subsection{Experiment 2: Discrimination of Grating Orientation}

The final experiment mirrored another classic (1981) study of human tactile perception~\cite{johnson}: a two-interval same-different psychophysical task on how the perception of grating orientation depends on its spatial period. We implemented a robot psychophysics version of the task on grating stimuli varying from smooth to coarse (0-5\,mm grating period; Figure \ref{tactile_images_presses}). The tactile robotic system made decisions on whether two gratings with the same period, presented sequentially, have the same or different (orthogonal) alignments. The perceptual system was build using convolutional neural network spatial decoders to estimate grating orientation from artificial SA-I and RA-I population responses, comprising two covert decisions within the sequential task. Overall, this task setup controlled the use of non-spatial mechanisms in the discrimination to isolate sensitivity to the grating period.  

Our initial investigation considered the spatial structure of artificial SA-I and RA-I tactile images representing the population responses. The grating period is visible in SA-I tactile images but not in RA-I tactile images, for grating periods of 3\,mm or more. Accordingly, the psychometric curves for grating discrimination in the same-different task using SA-I afferents had a similar sigmoid shape as for human participants (Figure 8B), increasing or decreasing from near random on the smoothest gratings to near-perfect discrimination on the coarsest gratings. Conversely, the task performance with RA-I afferents showed no improvement from random across the range, consistent with natural observations about their limited role in this task~\cite{johnson}. The similarity between human and artificial performance was quantified further with a standard psychometric measure, the just-noticeable diffeence, which was 2\,mm for artificial SA-I afferents and 1-7-2.1\,mm for human participants. Therefore, human performance is somewhat better on the finer gratings, as expected, but otherwise the artificial and human behaviour are qualitatively similar for grating orientation discrimination

\subsection{Conclusion}

In this paper, we saw that simple models of artificial SA-I and RA-I afferents using the TacTip soft biomimetic tactile sensor mimicked many properties of biological afferents in experiments designed for the study of natural touch. These tests included single afferent responses on complex ridged stimuli and perceptual decisions based on the afferent population responses during a grating-orientation task. The most significant discrepancy between the natural and artificial experimental results was that tactile skin from human and primate fingertips is considerably more sensitive to fine detail than the tactile sensor presented here. This discrepancy was not unexpected, because our 3d-printed artificial skin has a thicker and less conformable epidermis over a much lower density of artificial mechanoreceptors than human skin. We see the results of this paper as raising a challenge to the artificial tactile sensing community to fabricate a biomimetic tactile fingertip that reaches human tactile sensitivity while using transduction principles like those of natural mechanoreceptive afferents.

\subsection*{Data Accessibilty}

All code to process the data, train/test the deep learning models and generate result figures is available at https://github.com/nlepora/afferents-tactile-gratings-jrsi2022. Data are available at the University of Bristol data repository, data.bris, at https://doi.org/10.5523/bris.8jzw1c0jzjqj2jclybysvm0xj.

\subsection*{Author Contributions}
N.P. and N.L. planned the research. N.P. conducted the experiments and analysis, with help from T.G. on the psychophysics. N.P. and N.L. wrote the paper. N.L. did the revisions and final paper preparation.

\subsection*{Competing Interests}
We declare we have no competing interests.

\subsection*{Acknowledgements}
We thank the anonymous reviewers for their helpful and constructive advice. We also thank Stephen Redmond and Chris Kent for examining the PhD viva of N.P. upon which this paper is based. 

\subsection*{Funding}
This research was funded by a Leverhulme Research Leadership Award on `A biomimetic forebrain for robot touch' (RL-2016-39) and an EPSRC DTP PhD Scholarship for N.P.

\bibliographystyle{unsrt}
\bibliography{mybib}

\end{document}